\documentclass[letterpaper, 10 pt, conference, anonymous=true]{ieeeconf}  

\IEEEoverridecommandlockouts                            

\title{\LARGE \bf
Safe Explicable Policy Search
}

\author{
Akkamahadevi Hanni, Jonathan Montaño, and Yu Zhang
\thanks{*Akkamahadevi Hanni and Yu Zhang are with the School of Computing and Augmented Intelligence, Arizona State University, Tempe, USA
        {\tt\small ahanni@asu.edu} and {\tt\small Yu.Zhang.442@asu.edu}.}%
\thanks{Jonathan Montaño is with the School of Mathematical and Statistical Sciences, Arizona State University, Tempe, USA
        {\tt\small montano@asu.edu}.}%
}

\usepackage{pdfpages}
\usepackage{subfigure}
\usepackage{amsmath}
\usepackage{amsfonts}
\newtheorem{definition}{Definition}
\DeclareMathOperator{\argmax}{arg\,max}
\DeclareMathOperator{\argmin}{arg\,min}
\usepackage{pdfpages}
\usepackage{hyperref}
\usepackage{balance}
\usepackage{algorithm}
\usepackage{algpseudocode}
\newtheorem{theorem}{Theorem}

\begin{document}

\maketitle
\thispagestyle{empty}
\pagestyle{empty}

\begin{abstract}
When users work with AI agents, 
they form conscious or subconscious expectations of them.
Meeting user expectations is crucial for such agents to engage in successful interactions and teaming.
However, users may form expectations of an agent that differ from the agent's planned behaviors.
These differences lead to the consideration of two separate decision models in the planning process to generate explicable behaviors.
However, little has been done to incorporate safety considerations, especially in a learning setting.
We present \textit{Safe Explicable Policy Search (SEPS)}, which aims to provide a learning approach to explicable behavior generation while minimizing the safety risk, both during and after learning.
We formulate SEPS as a constrained optimization problem where the agent aims to maximize an explicability score subject to constraints on safety and a suboptimality criterion based on the agent's model. 
SEPS innovatively combines the capabilities of Constrained Policy Optimization and Explicable Policy Search to introduce the capability of generating safe explicable behaviors to domains with continuous state and action spaces, which is critical for robotic applications.
We evaluate SEPS in safety-gym environments and with a physical robot experiment to show its efficacy and relevance in human-AI teaming. 
\end{abstract}


\section{Introduction}
Recent developments in Artificial Intelligence (AI) have led to AI systems being widely deployed for user access. Users are increasingly involved in close interaction with these systems, forming a team-like relationship.  
In such cases, users form expectations of these ``teammates''. For successful teaming interactions, it is important that user expectations are met. Otherwise, it may lead to user confusion, poor teaming performance, and loss of trust \cite{Gunning_Aha_2019}. 

More specifically, users may not always be experts in AI systems, which may lead to them forming expectations that are different from the system's planned behaviors. These differences may be due to, for instance, different understandings of the task objective and/or domain dynamics during decision-making. 
Explicable planning \cite{Zhang2017PlanEA, gong2021explicable} addresses the planning challenges where the AI agent aims to generate more expected plans or policies under such model differences. However, when the agent focuses on meeting user expectations, 
it may lead to safety and/or efficiency issues.

\begin{figure}[t!]
    \subfigure[Motivating example]
  {\includegraphics[scale=0.185]{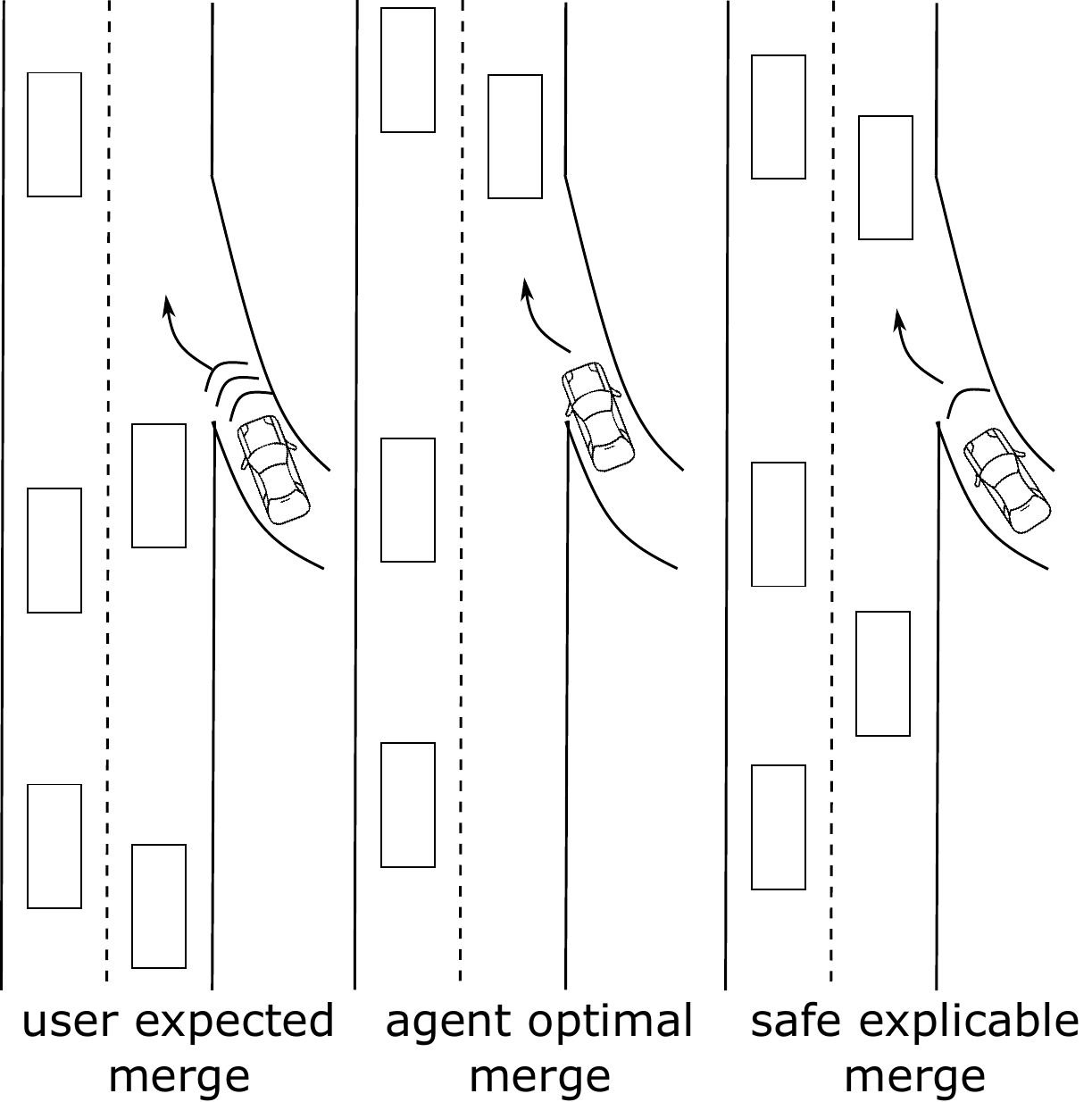}
  \label{fig:motivation}}
    \subfigure[Problem setting of SEP and the operation flow of SEPS]
  {\includegraphics[scale=0.35]{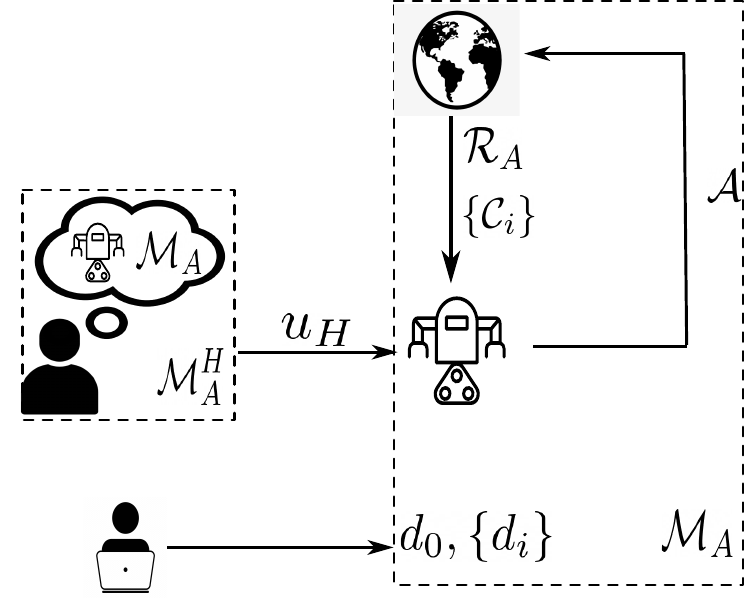}
  \label{fig:SEP-setting}}
    \caption{(a) The car merging scenario showing the user expected behavior (left), the optimal behavior (middle), and the safe explicable behavior (right), (b) The agent learns a surrogate reward function $u_H$ 
    that captures the user's expectations that are generated from her belief of the agent's model $\mathcal{M}_A$,  represented as $\mathcal{M}_A^H$. In a reinforcement learning setting,
    when the agent takes an action, it receives the agent's task reward $\mathcal{R}_A$, the safety costs associated with a set of cost functions, $\mathcal{C}_i$, and the assumed surrogate reward $u_H$ as user feedback. The agent approximates the expected returns for these costs and rewards while applying the quality $d_0$ and safety limits $\{d_i\}$ to ensure that its policy is safe and efficient. }
    \vskip-10pt
\end{figure}

Consider a motivating example, shown in Fig. \ref{fig:motivation}, where an autonomous driving car is to merge with highway traffic. The autonomous driving agent seeks to conserve energy so that it avoids hard acceleration and deceleration. However, such an objective can directly conflict with the user's expectation of minimizing the travel time. More specifically, while the optimal behavior of the agent is to slow down and wait for a large opening between the front and rear cars, the expected behavior is to merge as fast as it can, which can lead to safety risks. We can incorporate a safety criterion to maintain a minimum distance from the other cars to guarantee safety. However, an agent seeking explicable behaviors subject to the safety constraint may still opt for hard acceleration and deceleration with relatively small openings. A safe explicable behavior may involve alternating between soft deceleration and acceleration for an opportunity with a larger opening. The proposed Safe Explicable Policy Search (SEPS) aims to best match the user’s expectation within the agent’s safety and quality limits, thus avoiding unsafe and inefficient behaviors during and after learning.

A previous study \cite{hanni2024sep}, introduced the problem of Safe Explicable Planning (SEP) that aims to find user expected policies that are safe by considering a safety bound (provided by an expert) in the agent's decision making (Fig. \ref{fig:SEP-setting}). This work is limited to discrete states and actions, and to a planning setting where the models of the agent and the user (for generating expectations of the agent) are known. In reality, it is desirable for such methods to work in environments with continuous states and actions, and, more importantly, when the user's model is not directly available. Our work bridges the gap by proposing a method for finding safe and explicable policies in continuous stochastic environments in a reinforcement learning (RL) setting, which is especially valuable for robotic applications. 
It is worth noting that, in a planning setting, the solution is a final plan or policy that is safe. However, in a learning setting, the solution may require the agent to be safe even during the learning process.

Both safe behavior and explicable behavior generation, considered independently, can be addressed by the existing methods.  
For example, Constrained Markov Decision Process (CMDP) \cite{altman2021CMDP} and its solution techniques \cite{achiam2017constrained, pmlr-v119-wachi20a, Lyapunov_4fe51490}, etc., are effective in optimizing a task objective subject to safety constraints. 
The challenge in our problem is that the objective is defined over 
two different models with potentially different objectives and domain dynamics. This limits directly applying CMDP solutions. 
Similarly, a previous study on Explicable Policy Search (EPS) \cite{gong2021explicable} proposes learning a surrogate reward function from user feedback on the agent's behavior to inform learning.
It is more practical than learning both the reward and domain dynamics and hence can be applied when the user's model is unknown.  
However, EPS depends on the specification of a reconciliation parameter, and a weighting parameter for a few entropy terms, to combine an explicability score with the agent's objective.
It requires careful fine-tuning via trial and error due to the difficulty in relating these hyper-parameters to the importance of the objectives reconciled since they are at different scales (i.e., reward vs. entropy). These issues in existing methods limit their applicability to the generation of safe and explicable behaviors. 

To take advantage of both sides for safe explicable behavior generation, we first formulate our objective based on EPS since it allows us to convert the consideration of two models to the consideration of two linearly combined objectives. 
This is particularly beneficial since it enables us to focus policy search on a single model, reducing the new objective to a CMDP.  
Meanwhile, to relax the requirement of the hyper-parameters, 
we adapt and reformulate the objective by breaking it into an objective and a constraint, where the constraint can be dissolved into the CMDP formulation. We consider the popular Constrained Policy Optimization (CPO) \cite{achiam2017constrained} approach to solve the reduced CMDP problem. Our work considers a novel integration of EPS and CPO, while avoiding their limitations, to introduce safe explicable policy search. 
The remaining challenge lies in the derivation of an analytic solution for the resulting constrained optimization problem, for which we provide a solution for the most common case with two constraints.
Furthermore, we evaluate SEPS in different safety-gym environments and with a physical robot experiment. The results show that our approach is effective in searching for an explicable policy within the safety limits. Our approach also demonstrates its applicability to real-life scenarios in the robot experiment.



\section{Related Work}

With recent developments in AI, AI agents become accessible to various users, where they either work alongside or interact with them. This has motivated advancements in Explainable AI (XAI) with the aim of building AI agents whose behavior is explainable to users \cite{fox2017explainable,chakraborti2020emerging}. Our work is related to research that focuses on generating user-understandable behavior which is studied under different terminologies, viz., legibility \cite{dragan2013generating}, predictability~\cite{dragan2013LegPred}, transparency~\cite{macnally2018action}, explicability~\cite{Zhang2017PlanEA}, etc.
For a detailed review of these studies, refer to~\cite{chakraborti2019explicability}.

In the spectrum of Explainable AI literature, our work is most related to explicable planning \cite{Zhang2017PlanEA, gong2021explicable, Kulkarni2016ExplicableRP, hanni2021activeexp},  a framework for finding a user expected behavior while balancing the costs incurred to the agent. These studies propose different methods to measure similarity to user expectations but do not consider safety issues.
Other studies that consider human feedback in the agent's learning process, such as ~\cite{griffith2013policy, knox2009interactively, knox2012reinforcement}, are capable of learning user-preferred behavior in the presence of an expert designed RL objective. However, these methods assume the user's belief of the domain dynamics is the same as that of the agent, which can lead to learning undesired behaviors~\cite{gong2020you} when untrue. Furthermore, these approaches do not consider safety issues either. 

Safe Explicable Planning (SEP) extends explicable planning to support the specification of a safety bound~\cite{hanni2024sep}. However, it requires searching through the policy space, which is intractable in large, continuous MDPs.
Furthermore, SEP is limited to a planning setting in which the decision models are given.
In reality, such models may be difficult to provide a priori. The models can be learned individually, for example, \cite{reddy2018you} learned the domain dynamics and \cite{daniel2014active} learned the reward function, or jointly ~\cite{herman2016inverse, gong2020you}. However, for real-world problems, this can be challenging due to its complexity and high dimensionality. 
In this work, we consider the problem of finding a safe, explicable policy in a reinforcement learning setting. A prior work \cite{gong2021explicable} learns sufficient information about the user's model from user feedback, through a surrogate reward model.
In this work, we assume that the surrogate reward model is given. This allows us to focus on developing the methodology for SEPS. Furthermore, unlike SEP, in our formulation, we consider the general case of continuous state and action spaces.

In our work, we define the agent's decision model as a Constrained Markov Decision Process (CMDP) ~\cite{altman2021CMDP}, which allows us to specify the costs associated with safety separately from the agent's task objective.
CMDP is a popularly used formulation in Safe RL (refer to \cite{safeRLsurvey} for a review of Safe RL methods).
Another common way to handle safety in RL is to adopt a risk-aware formulation that includes a risk measure in the optimization criterion, which may take various forms, such as the exponential utility function ~\cite{howard1972risk, patek2001terminating}, a weighted sum of the return and risk functions ~\cite{gosavi2009reinforcement, geibel2005risk}, and probabilistic conditions ~\cite{kashima2007risk, morimura2010nonparametric}. 
In this work, we adopt the constrained formulation due to the ease of specification and its safety guarantees.
Several techniques have been proposed to solve a CMDP~\cite{achiam2017constrained, chow2018lyapunov, huang2022constrained, stooke2020responsive}. However, these methods cannot  be directly applied to solve SEPS due to its problem setting (refer to Fig. \ref{fig:SEP-setting}), where the optimization is under two different MDPs, specifically different transition models in MDPs.
We consider the Constrained Policy Optimization (CPO) \cite{achiam2017constrained} approach to solve CMDP due to its convergence and safety guarantees. Other recent studies ~\cite{yang2020projection, tessler2018reward, as2022constrained} that build on top of CPO can be easily applied to solve SEPS. 

\section{Preliminaries}
Our formulation is based on Markov Decision Processes (MDP). An MDP is represented by a tuple $\mathcal{M} = \langle \mathcal{S}, \mathcal{A}, \mathcal{T}, \mathcal{R}, \rho, \gamma \rangle$, where $\mathcal{S}$ is the set of states, $\mathcal{A}$ is the set of actions, $\mathcal{T}: \mathcal{S} \times \mathcal{A} \times \mathcal{S} \mapsto [0, 1]$ is the transition probability function, $\mathcal{R}: \mathcal{S} \times \mathcal{A} \times \mathcal{S} \mapsto \mathbb{R}$ is the reward function, $\rho:\Delta(\mathcal{S)}$ is the initial state distribution, and $\gamma$ is the discount factor. A stationary policy $\pi : \mathcal{S} \mapsto \Delta(\mathcal{A})$ is a map from states to probability distribution over actions, with $\pi(a|s)$ denoting the probability of selecting action $a$ in state $s$. We denote the set of all stationary policies by $\Pi$.
The performance (expected discounted return) of a policy w.r.t. a measure $\mathcal{R}$ under the transition dynamics $\mathcal{T}$ is typically given by $J_{\mathcal{R}, \mathcal{T}}(\pi) \dot{=} E_{\tau \sim \langle \pi, \mathcal{T}\rangle} [ \sum_{t=0}^{\infty} \gamma^t \mathcal{R} (s_t, a_t, s_{t+1}) ]$, where $\tau {\sim} \langle \pi, \mathcal{T} \rangle$ indicates that the distribution over trajectories depends on $\pi$ and $\mathcal{T}$ i.e., $s_0 {\sim}  \rho, a_t {\sim}  \pi(.|s_t), s_{t+1} {\sim}  \mathcal{T}(.|s_t, a_t)$. The probability of realizing a trajectory $\tau$ with policy $\pi$ is $p(\tau) {=} \rho(s_0) \prod_t \mathcal{T} (s_{t+1} | s_t, a_t) \pi ( a_t | s_t)$. 
Reinforcement learning algorithms (refer to \cite{Sutton1998}) seek to learn a policy that maximize the performance, i.e., $\pi^* = \argmax_{\pi \in \Pi} J_{\mathcal{R}, \mathcal{T}}(\pi)$.

\textbf{CMDP and Constrained Policy Optimization (CPO):} 
A CMDP is an MDP augmented with constraints that restrict the permissible policies for the MDP. Specifically, $\mathcal{M}$ is augmented by auxiliary cost functions $\{\mathcal{C}_i\}_{i=1:k}$ and limits $\{d_i\}_{i=1:k}$ where each $\mathcal{C}_i: \mathcal{S} \times \mathcal{A} \times \mathcal{S} \mapsto \mathbb{R}$ is a function, same as $\mathcal{R}$. 
The performance of a policy w.r.t. $\mathcal{C}_i$ is $J_{\mathcal{C}_i, \mathcal{T}}(\pi) = E_{\tau \sim \langle \pi, \mathcal{T} \rangle} [ \sum_{t=0}^{\infty} \gamma^t \mathcal{C}_i (s_t, a_t, s_{t+1}) ]$. Then, the set of all feasible stationary policies is denoted by $\Pi_{\mathcal{C}} \dot{=} \{ \pi {\in} \Pi : \forall{i} J_{\mathcal{C}_i, \mathcal{T}}(\pi) \leq d_i$\}. The objective of a CMDP is to learn a feasible policy that maximizes the performance $\pi^* = \argmax_{\pi \in \Pi_{\mathcal{C}}} J_{\mathcal{R}, \mathcal{T}}(\pi)$.

CPO \cite{achiam2017constrained} is a popular approach to solving CMDP.
CPO applies trust region policy optimization (TRPO) \cite{trpo-schulman15} to policy search updates in a CMDP.
The policy update rule is given by,
\begin{multline}
    \pi^{n+1} = \argmax_{\pi \in \Pi_\theta} \mathbb{E}_{\pi^n, \mathcal{T}} [ A_{\mathcal{R}}^{\pi^n} (s, a) ]\\
    s.t. J_{\mathcal{C}_i, \mathcal{T}}(\pi_n) + \frac{1}{1-\gamma} \mathbb{E}_{\pi^n, \mathcal{T}} [ A_{\mathcal{C}_i}^{\pi^n} (s, a) ] \leq d_i \hspace{10pt} i=1, ..., k \\
    \bar{\mathcal{D}}_{KL} (\pi || \pi_n) \leq \delta \hspace{60pt}
    \label{cpo}
\end{multline}
where, $A_{.}^\pi (s, a) \dot{=} Q_{.}^\pi (s, a) - V_{.}^\pi (s) $ is the advantage function given by the on-policy value function $V_{.}^\pi (s)$
and the on-policy action-value function $Q_{.}^\pi (s, a)$,
$\Pi_\theta \subset \Pi$ is a set of parameterized policies with parameters $\theta$, $\bar{\mathcal{D}}_{KL} (\pi || \pi_n) = \mathbb{E}_{s \sim \pi^n} [ \mathcal{D}_{KL} (\pi || \pi_n) [s]]$ is a measure of KL Divergence between the current policy $\pi_n$ and a potential new policy $\pi$, and $\delta > 0$ is the step size. The set of policies that satisfy the third constraint is called the trust region, i.e. $\{ \pi_\theta \in \Pi_\theta | \bar{\mathcal{D}}_{KL} (\pi || \pi_n) \leq \delta \}$. For performance guarantees and constraint satisfaction guarantees, see \cite{trpo-schulman15} and \cite{achiam2017constrained}, respectively. 

For continuous MDPs with high dimensional spaces, solving Eq. \eqref{cpo} exactly is intractable and requires an approximation. The CPO method proposes the following approximation
\begin{multline}
    \theta_{n+1} = \argmax_\theta g^T (\theta - \theta_n) \\
    s.t.\hspace{3pt} c_i + b_i^T  (\theta - \theta_n) \leq 0 \hspace{10pt} i=1, ..., k \\
    \frac{1}{2} (\theta - \theta_n)^T H (\theta - \theta_n) \leq \delta \hspace{80pt}
    \label{cpo-primal}
\end{multline}
where, $g$ is the gradient of the objective (explicability measure), $b_i$ is the gradient of the constraint $i$, $c_i \dot{=} J_{\mathcal{C}_i, \mathcal{T}}(\pi_n) - d_i$, and $H$ is the Hessian of the KL Divergence. 

\textbf{Explicable Policy Search (EPS):}
In EPS \cite{gong2021explicable}, two MDP models, the agent's model $M_A$ and the user's belief of the agent's model $M_A^H$, are considered at play. The two MDPs share the same $\mathcal{S}$, $\mathcal{A}$, $\rho$, and $\gamma$ but may differ in their domain dynamics ($\mathcal{T}_A$, $\mathcal{T}_A^H$) and their reward functions ($\mathcal{R}_A$, $\mathcal{R}_A^H$). 
In EPS, the aim is to search for a policy that maximizes 
a linear combination of two objectives, namely, the expected cumulative reward and a policy explicability score, weighted by a reconciliation factor ($\lambda$):
\begin{equation}
    \pi^* = \argmax_{\pi \in \Pi} J_{\mathcal{R}_A, \mathcal{T}_A}(\pi)
    + \lambda \hspace{2pt} \mathcal{E} (\pi, M_A, \pi_A^H, M_A^H).
    \label{eps}
\end{equation}

The explicability score $\mathcal{E}(.)$ is well motivated (refer to ~\cite{gong2021explicable}) and is defined as the negative KL-divergence of the trajectory distributions in the agent's model and the human's model denoted by $p_A(\tau)$ and $p_A^H(\tau)$, respectively. Formally, $\mathcal{E}(.) = - \mathcal{D}_{KL} (p_A(\tau) || p_A^H(\tau))$ ${=} - \mathbb{E}_{p_A} [\sum_t \log \mathcal{T}_A (s_{t+1} | s_t, a_t) + \log \pi_A ( a_t | s_t) - \log \mathcal{T}_A^H (s_{t+1} | s_t, a_t) - \log \pi_A^H ( a_t | s_t) ] + C$. Essentially, the agent that maximizes $\mathcal{E}(.)$ would learn to replicate (as closely as possible) the human's expected behavior of the agent. 

Estimating the parameters in $p_A^H(\tau)$ (i.e. $\mathcal{T}_A^H$ and $\pi_A^H$), individually, is expensive. 
Instead, the authors propose to learn a surrogate reward function $u_H$ via a preference-based learning framework \cite{christiano2017deep, knox2022models} by presenting pairs of trajectories and asking human subjects which trajectory is expected. 
The goal of learning $u_H$ is to obtain feedback on the user's assessment of \textit{explicableness} 
of the agent's behavior, which is evaluated under the user's belief (i.e., $\mathcal{M}_A^H$).
The learned reward function $u_H$ correlates with the distribution of human expected trajectories i.e., $u_H(s_t, a_t) =\log  \mathcal{T}_A^H (s_{t+1} | s_t, a_t) + \log \pi_A^H ( a_t | s_t) + C_1$. 
Assuming a parametrized agent policy $\pi_\theta$, the surrogate reward and other log terms in the explicability score effectively reshape the reward function $\mathcal{R}_A$, and Eq. \eqref{eps} is approximated as follows 
\begin{multline}
    \theta^* \dot{=} \argmax_{\pi_\theta} 
    \mathbb{E}_{p_A} 
    \Bigl[ \sum_t \gamma^t 
    (\mathcal{R}_A (s_t, a_t) \\
    + \lambda (u_H (s_t, a_t) + \mathcal{H}_{\mathcal{T}_A} (s_{t+1} | s_t, a_t) + \mathcal{H}_{\pi_\theta} (a_t | s_t))) \Bigr],
    \label{eps2}
\end{multline}
where $\mathcal{H}_{\pi_\theta}$ and $\mathcal{H}_{\mathcal{T}_A}$ are the entropies of the agent's policy and domain dynamics (i.e., the true environment dynamics), respectively.


\section{Problem Formulation}

To formulate safe explicable policy search (SEPS), similar to EPS, the user's belief of the agent's model is modeled as an MDP i.e., $\mathcal{M}_A^H = \langle \mathcal{S}, \mathcal{A}, \mathcal{T}_A^H, \mathcal{R}_A^H, \rho, \gamma \rangle$;
furthermore,
the agent's decision model is modeled as a CMDP i.e., $\mathcal{M}_A = \langle \mathcal{S}, \mathcal{A}, \mathcal{T}_A, \mathcal{R}_A, \{\mathcal{C}_i\}_{i=1:k}, \{d_i\}_{i=0:k}, \rho, \gamma \rangle$.
To focus on policy search, we assume access to the surrogate reward function $u_H$ in EPS.

\subsection{Rationale}
Explicability and safety are orthogonal issues: an explicable policy may place too much focus on user understandability, resulting in unsafe behaviors when the user is simply unaware of the safety risks; on the other hand, an explicable policy may align well with safety if the user understands the risks but is risk-averse, resulting in overly conservative behaviors that negatively impact task efficiency. Hence, it is imperative for explicable behavior generation to be guarded by both safety constraints and efficiency criteria.
Hence, it is imperative for explicable behavior generation to be guarded by safety constraints. However, we will explain next that a few alternatives would not work well. 


\subsubsection{Multi-objective optimization with linear combination}
One idea is to optimize an objective that balances safety costs with rewards and explicability. 
The corresponding optimization problem can then be defined as maximizing $\mathcal{R}' = \mathcal{R} - \omega_1 \mathcal{C} + \omega_2 \mathcal{E}$, where $\omega_1$ and $\omega_2$ are the respective scaling parameters. 

\subsubsection{Constrained EPS}
Alternatively, we can consider EPS subject to safety constraints as follows:
\begin{align}
    \pi^* = \argmax_{\pi \in \Pi_d} &J_{\mathcal{R}_A, \mathcal{T}_A}(\pi)
    + \lambda \hspace{2pt} \mathcal{E} (\pi, \mathcal{M}_A, \pi_A^H, \mathcal{M}_A^H) \notag \\
    s.t. \hspace{3pt} &J_{\mathcal{C}_i, \mathcal{T}_A}(\pi) \leq d_i \hspace{10pt} i=1, ...., k.
    \label{alt-seps}
\end{align}

The above two alternative formulations considered share two related limitations: 1) determining an appropriate reconciliation/weight factors can be difficult.  Designing a single reward that simultaneously drives task completion, meets user expectations, and discourages unsafe behavior is challenging.
It requires carefully scaling each component to provide a meaningful learning signal at each step, preventing issues like reward hacking.
For example, in CPO [6], the results of optimizing $\mathcal{R} - \omega_1 \mathcal{C}$ demonstrate a high sensitivity to the cost coefficient $\omega_1$, highlighting the difficulty of crafting an effective overloaded reward function. Prior research [29], also emphasizes the importance of separating the task objective from safety constraints, as in CMDPs \cite{altman2021CMDP, achiam2017constrained, pmlr-v119-wachi20a, chow2018lyapunov}, to enable the agent to learn meaningful behavior.
2) since the task objective is maximized in combination with explicability, for $\lambda > 0$, there is no guarantee of task performance or even achievement, since the agent may be able to increase the explicability score to ``compensate" for the reduction in task performance. 

\subsection{Safe Explicable Policy Search (SEPS)}


In SEPS, we observe that the two components of EPS can be separated to remove the reconciliation factor and let the agent reward be absorbed into the safe constraints in CMDP, with the limit $d_0$. Even though this would introduce a constraint parameter, it has an intuitive appeal: the suboptimality level of the behavior, which can be used to ensure the task performance/achievement as desired.  Hence, such a parameter can be set more straightforwardly by the engineer than the semantically confusing reconciliation factor.  

A remaining challenge is the additional entropy terms in $\mathcal{E}$. Since the surrogate reward $u_H$ in EPS is learned to reflect
$\mathcal{H}_{\mathcal{T}_A^H}$ and $\mathcal{H}_{\pi_A^H}$, combining it with the remaining entropy terms ($\mathcal{H}_{\mathcal{T}_A}$ and $\mathcal{H}_{\pi_\theta}$) requires additional scaling parameters. 
In SEPS, we simplify by removing these terms. 
This means that instead of maximizing the negative KL divergence between the trajectory distributions, we maximize the probability of the most expected trajectory that the agent can generate.
This leads to an approximation of the explicability objective that encourages a narrower distribution than the original objective.  

\begin{definition}
    Safe Explicable Policy Search (SEPS) is the problem of searching for a policy that maximizes the probability of the user's expected trajectory, captured by $u_H$, subject to the constraints on the task objective $\mathcal{R}_A$ and the costs $\mathcal{C}_i$ associated with safety in the agent's model, or formally,
\end{definition}
\begin{align}
        \pi^* &\dot{=} \argmax_{\pi \in \Pi_d} J_{u_H, \mathcal{T}_A}(\pi) \notag \\
        s.t. \hspace{3pt} &\color{black}{C_0: \,\,\,\,} J_{\mathcal{R}_A, \mathcal{T}_A}(\pi) \geq d_0 \notag \\
        &\color{black}{C_i: \,\,\,\,} J_{\mathcal{C}_i, \mathcal{T}_A}(\pi) \leq d_i \hspace{10pt} i=1, ..., k.    
    \label{seps}
    \end{align}

The main objective here is to maximize $u_H$ under the true domain dynamics $\mathcal{T}_A$, which in turn maximizes the approximate explicability score.
Although the SEPS problem in Eq. \eqref{seps} appears similar to the SEP \cite{hanni2024sep} problem below
\begin{align}
    \pi^* = \argmax_{\pi \in \Pi_d} &J_{\mathcal{R}_A^H, \mathcal{T}_A^H}(\pi) \notag \\
    s.t. \hspace{3pt} J_{\mathcal{R}_A, \mathcal{T}_A}(\pi) &\geq d_0,
    \label{sep}
\end{align}
there are a few critical differences: 1) in SEPS, the constraint on the agent's reward
or task performance, whereas it was associated with the safety constraint in [4]. It requires the safety constraint to be correlated with task performance, which is not always feasible;
2) in \cite{hanni2024sep}, $\mathcal{M}_A^H$ is assumed to be given whereas it was replaced here by the surrogate reward function that can be learned, which is evaluated under the true domain dynamics. Such a subtle but important difference allows us to apply learning methods to solve the problem instead of replying on planning; 3) formulating SEPS as a CMDP allows us to leverage CMDP solutions to handle multiple constraints in continuous domains, instead of as a complex search problem under two different models in \cite{hanni2024sep}. 


\section{Search Method}

When the MDP models are known, as in Eq. \eqref{sep}, SEP \cite{hanni2024sep} proposes searching through the policy space while evaluating policies in the agent's model for feasibility and in the user's model for explicablity. The SEP algorithms iterate over all unique deterministic policies (formed by iterating over all states and actions) to find an optimal safe, explicable policy, which is impractical in continuous MDPs.
The SEPS problem in Eq. \eqref{seps} reduces to a CMDP, allowing us to use its policy search techniques. 

We choose the CPO \cite{achiam2017constrained} algorithm to solve SEPS as it ensures safe behavior even during the learning process. However, the CPO algorithm is limited to solving an objective with a single constraint and does not provide the analytic solution for more than one constraint. 
As the number of constraints increases, deriving feasible bounds for the solution becomes challenging because of the intersection of feasible regions of every constraint. 
In this work, we adopt the CPO solution technique, but derive a new analytical solution for SEPS that includes at least two constraints.

\textcolor{black}{Applying the CPO update rule to SEPS in Eq. \eqref{seps} will yield an expression similar to that in Eq. \eqref{cpo}. An approximation to that update rule (Eq. \eqref{cpo-primal} as in CPO) is given by the primal below:}
\begin{align}
    \theta_{n+1} = \argmax_\theta g^T &(\theta - \theta_n ) \notag \\
    s.t.\hspace{3pt} c_0 - b_0^T  (\theta - \theta_n) &\leq 0 \notag \\
    c_1 + b_1^T  (\theta - \theta_n) &\leq 0 \notag \\
    \frac{1}{2} (\theta - \theta_n)^T H (\theta - \theta_n) &\leq \delta 
    \label{feas-primal}
\end{align}
where, $b_0$ is the gradient of the constraint on agent's task objective, and $c_0 \dot{=} - J_{\mathcal{R}_A, \mathcal{T}_A}(\pi_n) + d_0$. $b_1$ and $c_1$ are as defined in Eq. \eqref{cpo-primal}.


     

The optimization problem mentioned above is convex. It is worth noting that for policies with high-dimensional parameter spaces like neural networks, the primal problem (even with few constraints) is impractical to solve using a convex program solver due to the computation complexity of the Hessian matrix, which motivates the need for an efficient analytical solution.

\textbf{Feasible case:}
We consider the feasible case when the current policy $\pi_{\theta_n}$ satisfies all constraints. In this case, it is easier to solve the dual of Eq. \eqref{feas-primal}, given by
\begin{equation}
    \max_{\substack{\lambda \geq 0 \\ \nu_0 \geq 0 \\ \nu_1 \geq 0 }} g^T x + \lambda ( \frac{1}{2} x^T H x - \delta ) + \nu_0 ( b_0^T x + c_0 ) + \nu_1 ( b_1^T x + c_1 )
    \label{fescase}
\end{equation}
where $\lambda$, $\nu_0$, and $\nu_1$ are Lagrangian variables.
The solution to the feasible case is given by Theorem 1 in Appendix A of the extended version of this paper \cite{SEPS-full-version}.
To determine the optimal values of the Lagrangian variables, we need to consider the intersection of the trust region with the two linear constraints. Specifically, the optimal value of $\lambda$ depends on the following four cases, viz., 1) both linear constraints are active ($\nu_0 > 0$ and $\nu_1 > 0$), 2) only the first linear constraint is active ($\nu_0 > 0$ and $\nu_1 = 0$): which means that the trust region lies well within the feasible region of the second constraint and can be safely ignored, 3) only the second linear constraint is active ($\nu_0 = 0$ and $\nu_1 > 0$), and 4) both linear constraints are inactive ($\nu_0 = 0$ and $\nu_1 = 0$): which means that the optimization depends solely on the objective ($g^T x$).
In brief, if $\lambda^*$, $\nu_0^*$, and $\nu_1^*$ are optimal solutions to the dual above, then,
the update rule for the feasible case is 
\begin{equation}
    \theta_{n+1}^* = \theta_n + \frac{1}{\lambda^*} H^{-1} \left( g + \nu_0^* b_0 - \nu_1^*b_1 \right) 
    \label{feas-update}
\end{equation}

\textbf{Infeasible case:}
We consider the infeasible case when the current policy $\pi_{\theta_n}$ violates one or both linear constraints. This could be the result of a bad step taken previously (occurs due to errors in approximation). The system can recover by updating the policy parameters to strictly reduce the constraint violation. In other words, the system must focus on satisfying the constraint(s) and ignore optimizing the objective ($g^T x$), temporarily. 
In this case, depending on the linear constraint that is violated in Eq. \eqref{feas-primal}, the update rule is found by solving the primal below 
\begin{multline}
    \theta_{n+1} = \argmin_\theta \hspace{5pt} c_m + b_m^T  (\theta - \theta_n) \hspace{15pt} \text{where, } m \in \{0,1\} \\
    s.t. \,\,\,\,  \frac{1}{2} (\theta - \theta_n)^T H (\theta - \theta_n) \leq \delta, \hspace{50pt}
    \label{inf-primal}
\end{multline}
and its corresponding dual is given by
\begin{equation}
    \max_{\phi \geq 0} -\frac{b_m^T H^{-1} b_m}{2\phi} + c_m - \phi \delta.
    \label{infcase}
\end{equation}
where $\phi$ is a Lagrangian variable.
The solution to the feasible case is given by Theorem 2 in Appendix A of the extended version of this paper \cite{SEPS-full-version}.

In brief, if $\phi^*$ is the solution to the dual, then the update rule for the infeasible case is 
\begin{equation}
    \theta_{n+1}^* = \theta_n - \frac{1}{\phi^*} H^{-1} b_m
    \label{inf-update}
\end{equation}

When multiple constraints are violated, we solve Eq. \eqref{inf-primal} for each constraint that is violated and linearly combine the solutions in Eq. \eqref{inf-update}. This adjusts the policy towards the feasible region w.r.t. all constraints in a single step. Alternatively, one could update the policy with respect to a single constraint at a time.

\section{Evaluation}

We evaluated our method in continuous Safety-Gym environments \cite{ji2023safety} and through a physical robot experiment, designing scenarios that included explicit risk elements. Across all evaluations, we assumed a learned surrogate reward function that captures sufficient information about user expectations in $\mathcal{M}_A^H$. Our goal is to 1) validate the ability of our method to find safe, efficient, and explicable policies that meet user expectations, 2) analyze the behaviors generated by our method relative to state-of-the-art and baseline approaches, and 3) demonstrate the applicability of SEPS in real-world scenarios.

\subsection{Safety}
\begin{figure}[ht!]
    \centering
    \subfigure[PointGoal]
    {\includegraphics[scale=0.155]{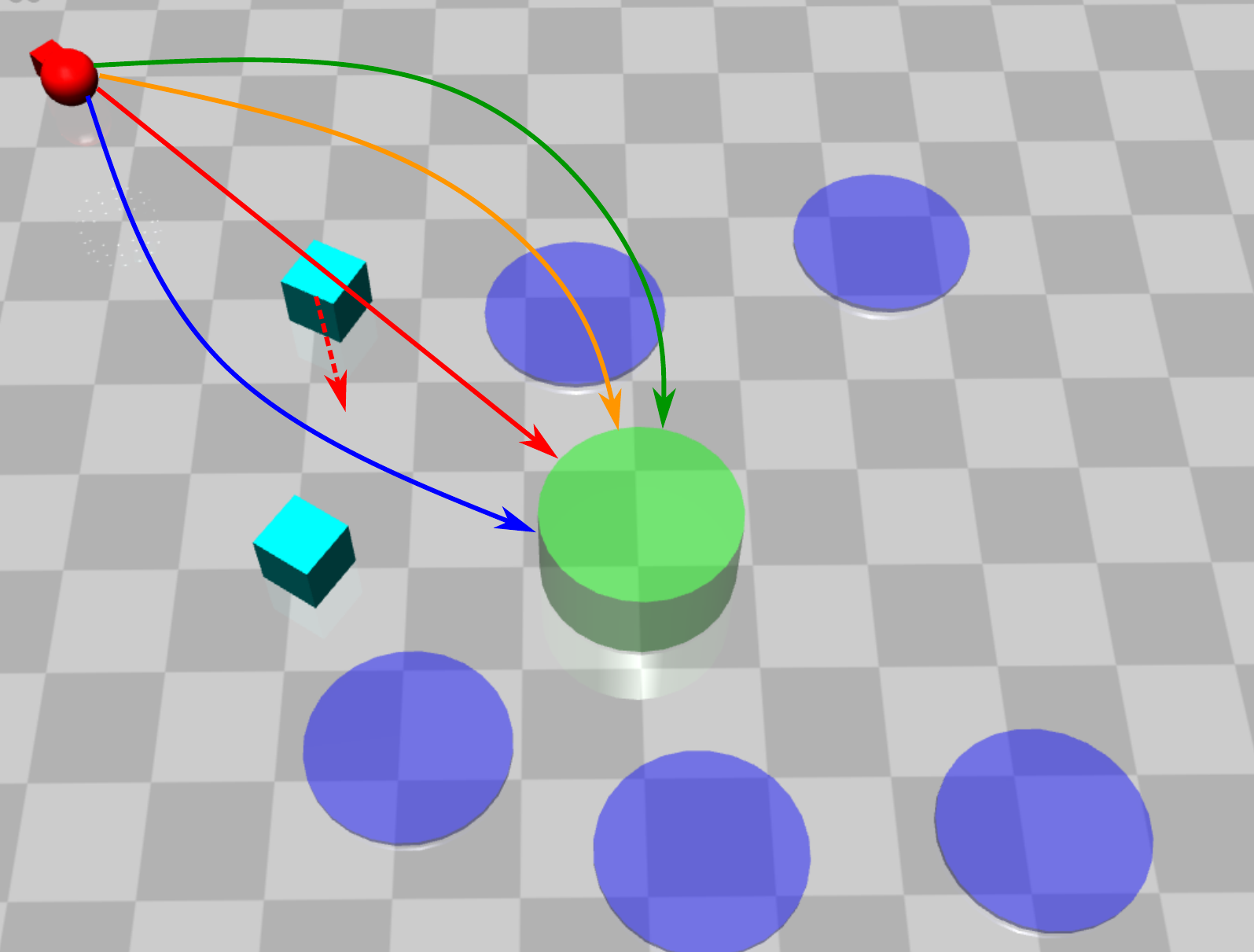}
    \label{fig:exp1-behavior}}
    \hspace{0.5pt}
    \subfigure[PointButton]
    {\includegraphics[scale=0.161]{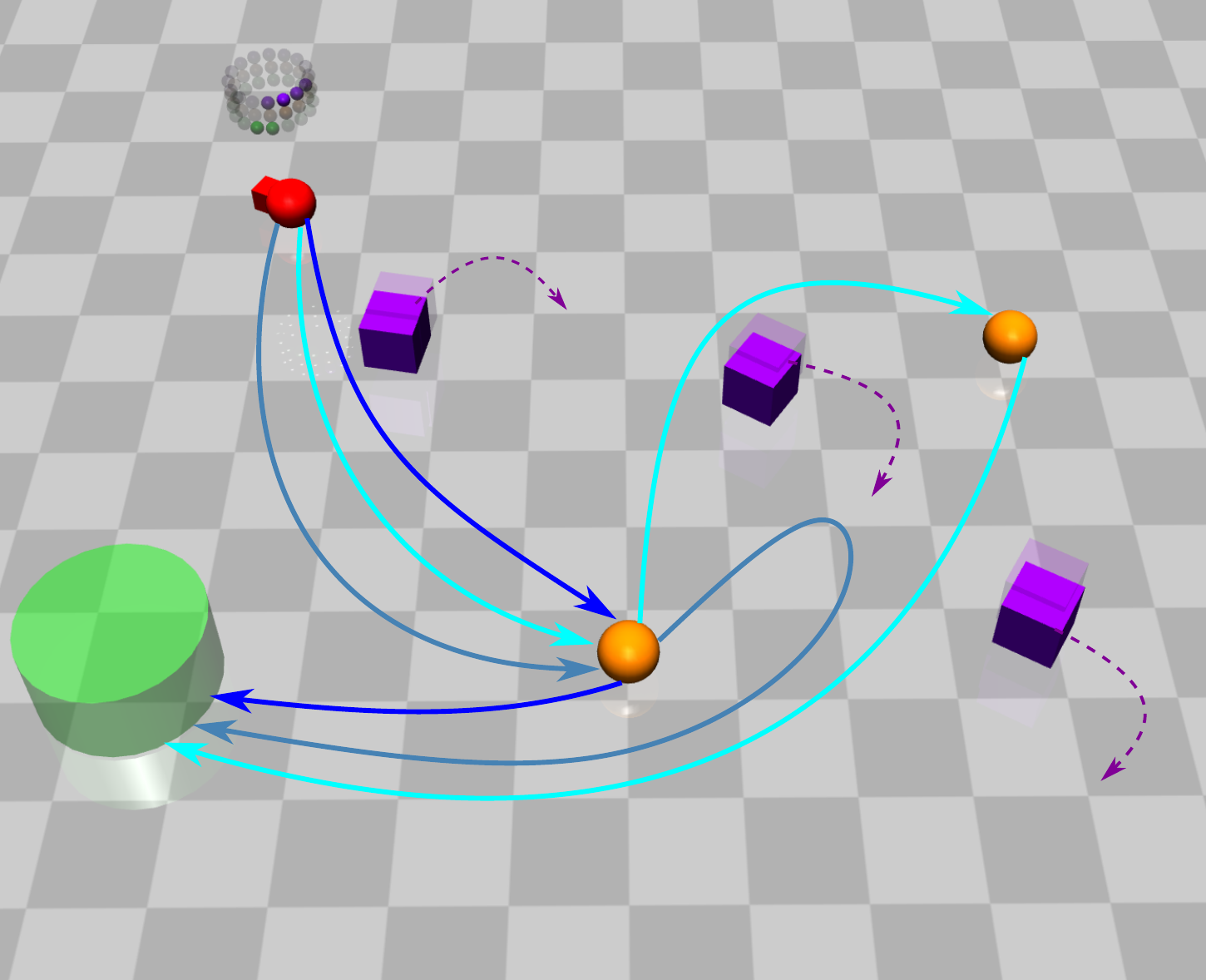}
    \label{fig:exp2-behavior}}
    \caption{Lines show resulting trajectories of different methods in the two simulation domains; (a) SEPS - blue, AGT - red, HUM - orange, EPS - green, (b) SEPS - blue, SEPS$-C_0$ - light blue, SEPS$+\mathcal{R}_A{-}C_0$ - medium blue.
    }
    \label{fig:exp12-behaviors}
\end{figure}

In our first evaluation, our goal is to answer `What are the benefits of safety constraints in explicable planning?' and `Can SEPS identify an explicable policy under safety constraints?'. We use the Safety-Gym PointGoal task (shown in Fig. \ref{fig:exp1-behavior}) where we assume user expectations conflict with the agent's safety requirements. In $\mathcal{M}_A$, the agent must reach the goal (green) without entering hazardous regions (dark blue). The reward $\mathcal{R}_A$ is positive for progress toward the goal and negative for moving away, and $\mathcal{C}_1$ penalizes entry into hazardous regions.
In contrast, according to user expectations ($u_H$), the agent can traverse hazardous regions as the user is unaware of their risks but must avoid the light blue boxes as the user regards them fragile.
\begin{figure*}[!ht]
\centering
\subfigure[Returns under $u_H$]{\includegraphics[scale=0.275]{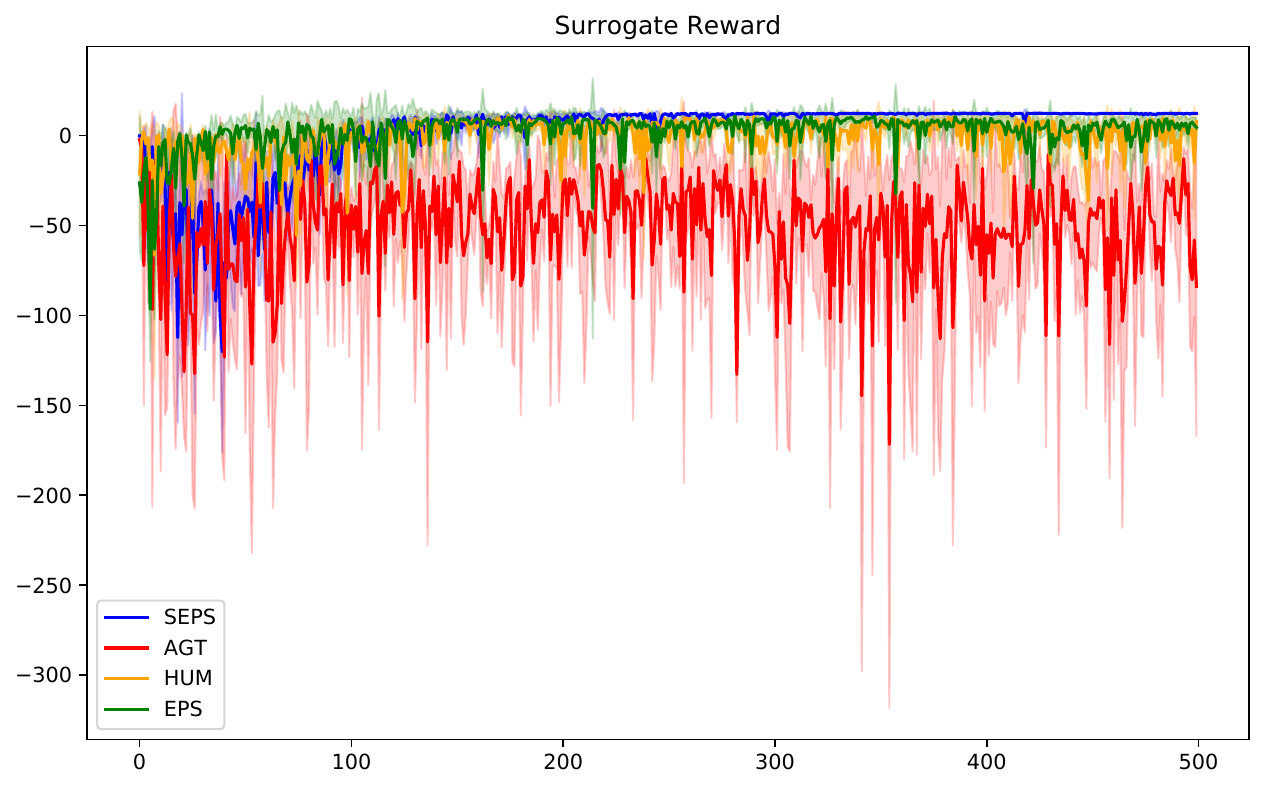}\label{fig:exp1.a}}
\subfigure[Returns under $\mathcal{R}_A$]{\includegraphics[scale=0.275]{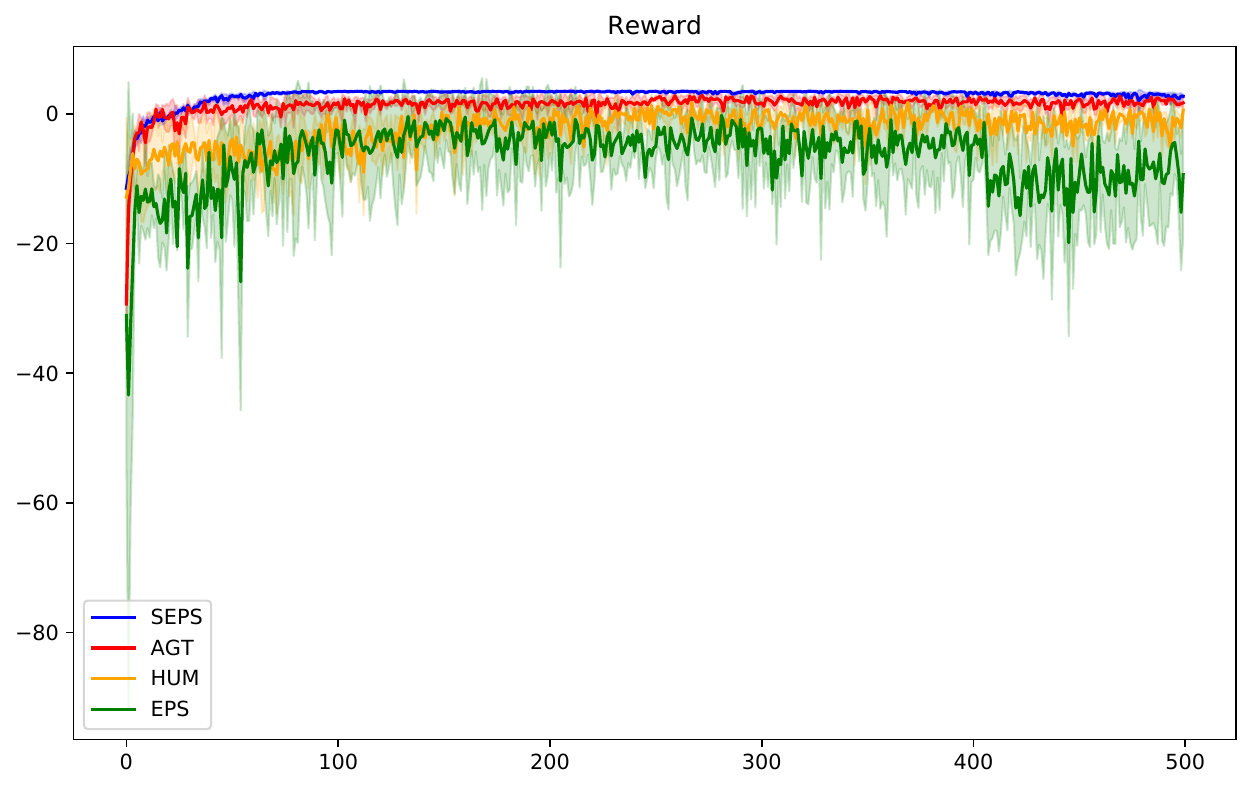}\label{fig:exp1.b}}
\subfigure[Returns under $\mathcal{C}_1$]{\includegraphics[scale=0.275]{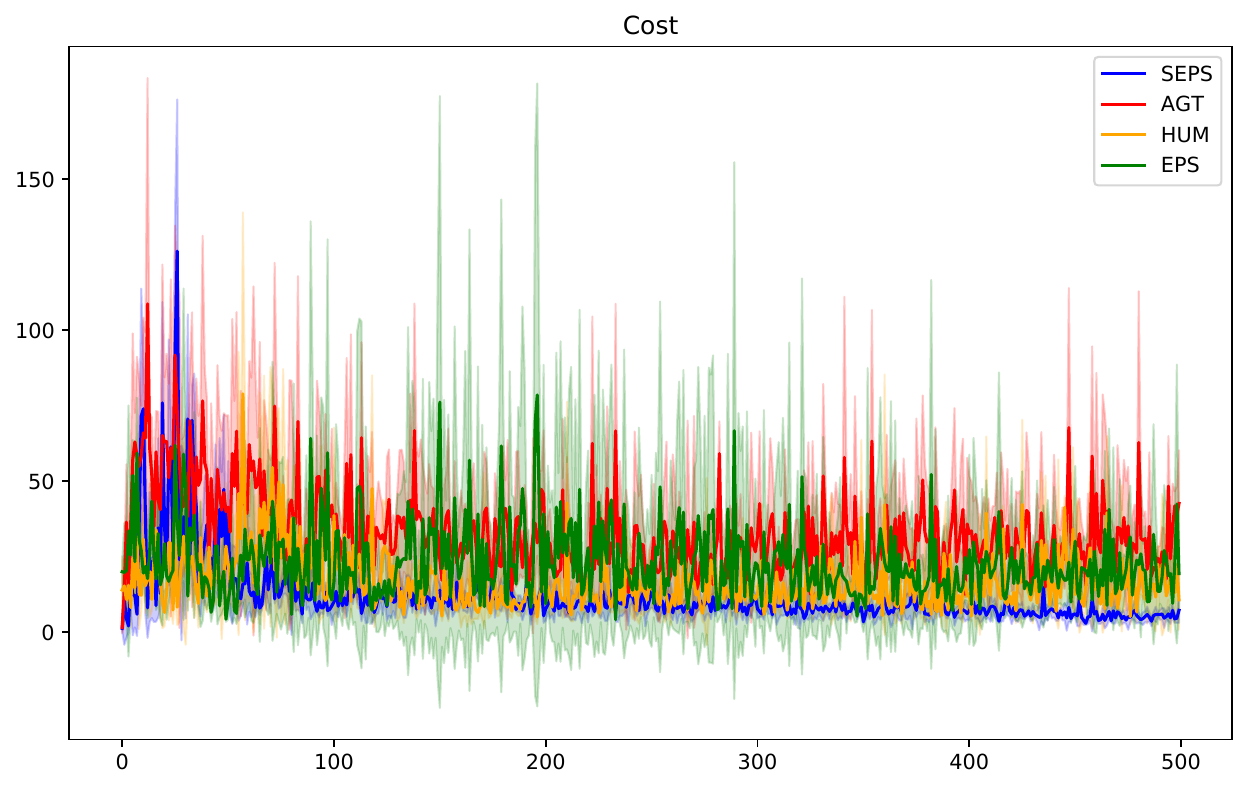}\label{fig:exp1.c}}
\caption{Average performance over three runs for SEPS and baselines in PointGoal; the x-axis is training epoch.}
\label{fig:exp1}
\vskip-10pt
\end{figure*}

We evaluated SEPS against three baselines: AGT (optimal agent behavior), HUM (human expected behavior), and EPS. AGT optimizes $\mathcal{R}_A$, HUM optimizes $u_H$, and EPS optimizes the linear combination $\mathcal{R}_A + \lambda u_H$ (see the Preliminary Section). All baselines were trained with PPO \cite{schulman2017proximal}. Here, the need for safety constraints in explicable planning is highlighted, as the baselines do not incorporate them. We compared the resulting behaviors and measured the returns under the surrogate reward, the agent’s task reward, and the cost function.

\noindent\textbf{Results and Discussion.}
Figure \ref{fig:exp1-behavior} illustrates the behaviors generated. AGT reached the goal via the shortest path by moving one of the boxes, contrary to user expectations. HUM and EPS avoided the boxes, but entered hazardous regions due to a lack of safety constraints. In contrast, SEPS efficiently reached the goal while avoiding both boxes and hazardous regions.

These behaviors aligned with the quantitative results in Fig. \ref{fig:exp1}. The primary finding is that all baselines violated safety constraints, as reflected by returns in $\mathcal{C}_1$ (Fig. \ref{fig:exp1.c}), where values farther from zero indicate greater violations. AGT, HUM, and EPS showed higher violations as their objectives ignore safety specifications. AGT also performed poorly on user expectations (Fig. \ref{fig:exp1.a}), prioritizing the shortest path even if it involves moving boxes. HUM achieved high returns in $\mathcal{R}_A$ (Fig. \ref{fig:exp1.b}) as it does not deviate much from the goal. EPS, despite using the best reconciliation factor ($\lambda=2$), showed instability in learning $\mathcal{R}_A$ (Fig. \ref{fig:exp1.b}) due to its linear combination objective. SEPS outperformed the baselines on the agent’s task (Fig. \ref{fig:exp1.a}) and user expectations (Fig. \ref{fig:exp1.b}) while satisfying safety requirements (Fig. \ref{fig:exp1.c}).

\subsection{Ablation Study}
In our second evaluation,  our goal is to answer `Can safe explicable policy search be formulated differently?' or, equivalently, `What is the benefit of the SEPS formulation?'. We used the Safety-Gym PointButton task (Fig. \ref{fig:exp2-behavior}), where the user’s model aligns with the agent’s safety requirements but differs from its task objective.

\begin{figure*}[!ht]
\centering
\subfigure[Returns under $u_H$]{\includegraphics[scale=0.275]{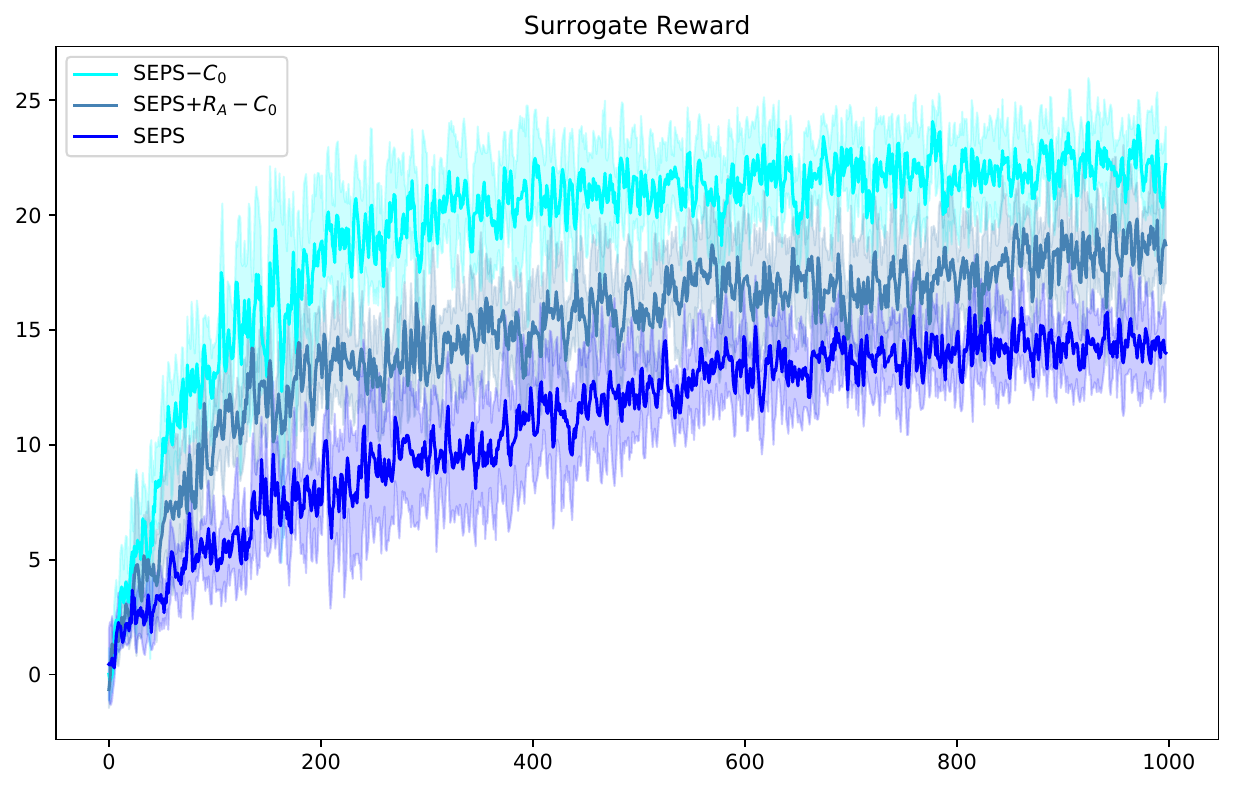}\label{fig:exp2.a}}
\subfigure[Returns under $\mathcal{R}_A$]{\includegraphics[scale=0.275]{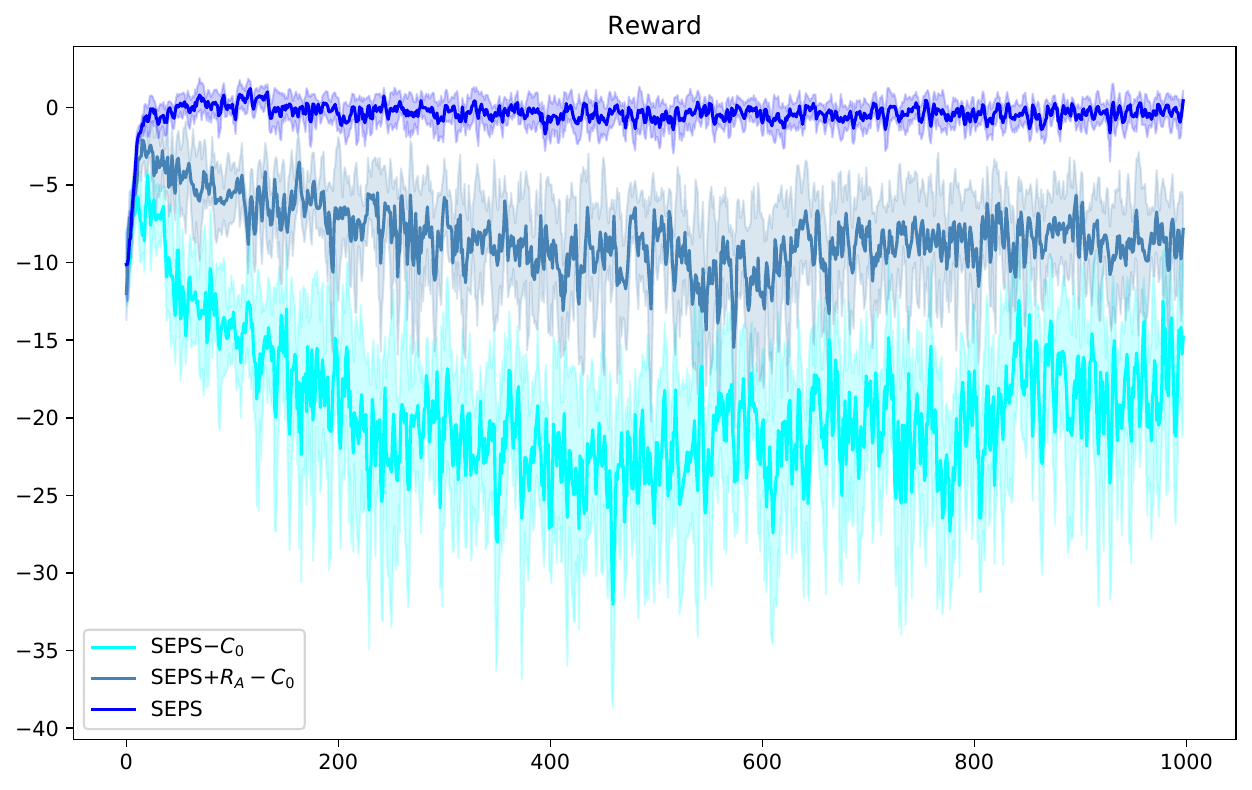}\label{fig:exp2.b}}
\subfigure[Returns under $\mathcal{C}_1$]{\includegraphics[scale=0.275]{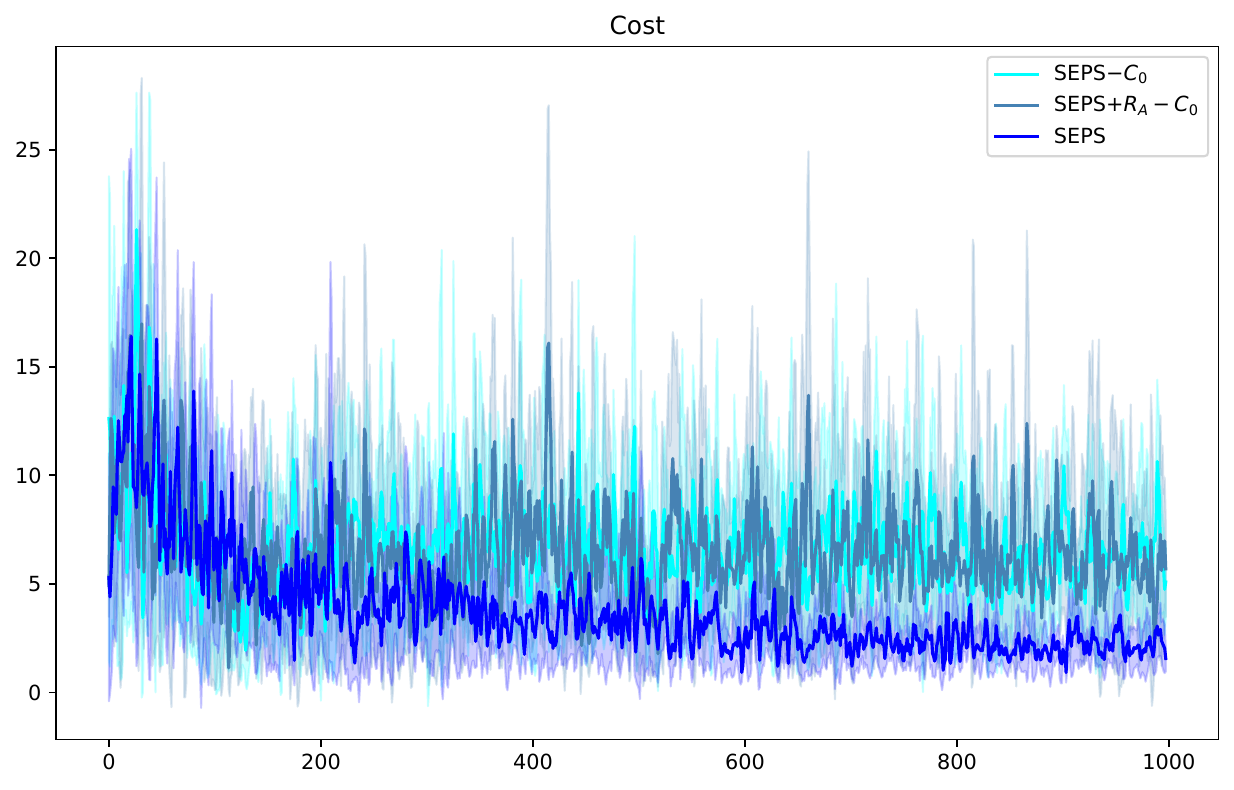}\label{fig:exp2.c}}
\caption{Average performance over three runs for SEPS and baselines in PointButton task over random seeds; the x-axis is training epoch.}
\label{fig:exp2}
\vskip-15pt
\end{figure*}

In $\mathcal{M}_A$, the agent must reach the goal (green) without colliding with gremlins (purple), which move in fixed patterns, and may optionally press buttons (orange). The reward $\mathcal{R}_A$ assigns positive values for reaching the goal and pressing buttons and negative values for moving away from the goal, while the cost $\mathcal{C}_1$ penalizes collisions with gremlins. According to user expectations ($u_H$), the agent must press both buttons before reaching the goal. Thus, $u_H$ mirrors $\mathcal{R}_A$ except that deviations from the goal do not incur a negative value.

We compared the formulation of SEPS with two ablation baselines: SEPS$-C_0$, which removes the constraint $C_0$ (refer to Eq. \eqref{seps}), and SEPS$+\mathcal{R}_A{-}C_0$, which linearly combines the SEPS objective with $\mathcal{R}_A$ while omitting $C_0$. Both baselines were trained using CPO \cite{achiam2017constrained}, as they involve a single constraint. We evaluated them by analyzing the generated behaviors and returns under the surrogate reward, the agent’s task reward, and the cost function.

\noindent\textbf{Results and Discussion.}
Figure \ref{fig:exp2-behavior} shows the generated behaviors. All baselines avoided gremlins due to safety constraints. SEPS$-C_0$ optimized user expectations, pressing both buttons before reaching the goal. SEPS pressed only the closer button to avoid large penalties for deviations, thereby satisfying the suboptimality criterion. SEPS$+\mathcal{R}_A{-}C_0$ also pressed the closer button but exhibited random deviations (Fig. \ref{fig:exp2-behavior}) due to optimizing linearly combined objectives.

These behaviors align with the quantitative results in Fig. \ref{fig:exp2}. SEPS$-C_0$ performed poorly under $\mathcal{R}_A$ as it focuses solely on $u_H$. SEPS$+\mathcal{R}_A{-}C_0$, despite optimizing both $u_H$ and $\mathcal{R}_A$ (with $\lambda{=}3$), also underperformed in $\mathcal{R}_A$ because it lacked constraints to enforce efficiency. In contrast, SEPS enforces the suboptimality criterion and as a result it guarantees efficiency in the agent’s task objective.

In SEPS, we set $d_0=0.0$ and $d_1=2.5$. These limits have intuitive semantic interpretations. For example, the agent can deviate from the goal provided that its expected return in $\mathcal{R}_A$ remains above $0.0$, making them easier to specify than the hyperparameters in EPS. Unlike SEPS, all baselines impose only a single safety constraint, whereas handling multiple constraints can be more challenging due to potential conflicts. Even under such conditions, SEPS remained close to $\mathcal{M}_A$ in both $\mathcal{R}_A$ and $\mathcal{C}_1$, demonstrating that it can remain explicable while meeting both performance and safety requirements.

\subsection{Physical Robot Experiment}
\begin{figure}[ht!]
    \centering
    \subfigure[$\tau_{\text{AGT}}$]
    {\includegraphics[scale=0.35]{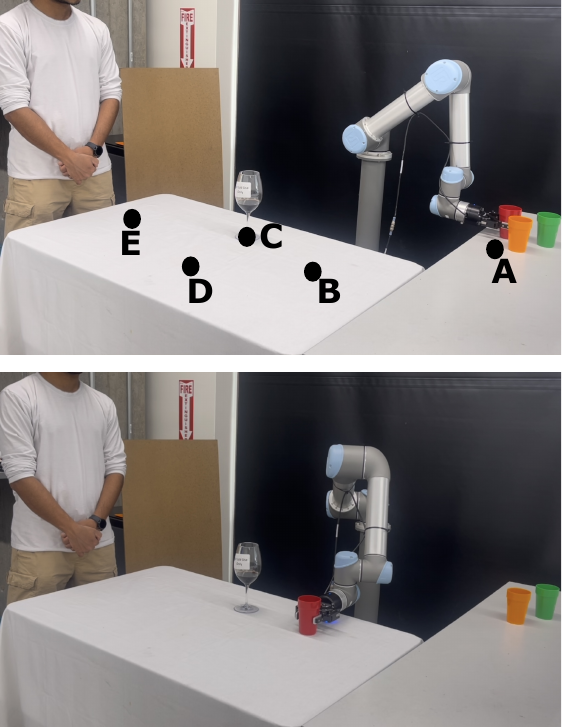}
    \label{fig:ur5-mr}}
    \hspace{20pt}
    \subfigure[$\tau_{\text{HUM}}$]
    {\includegraphics[scale=0.35]{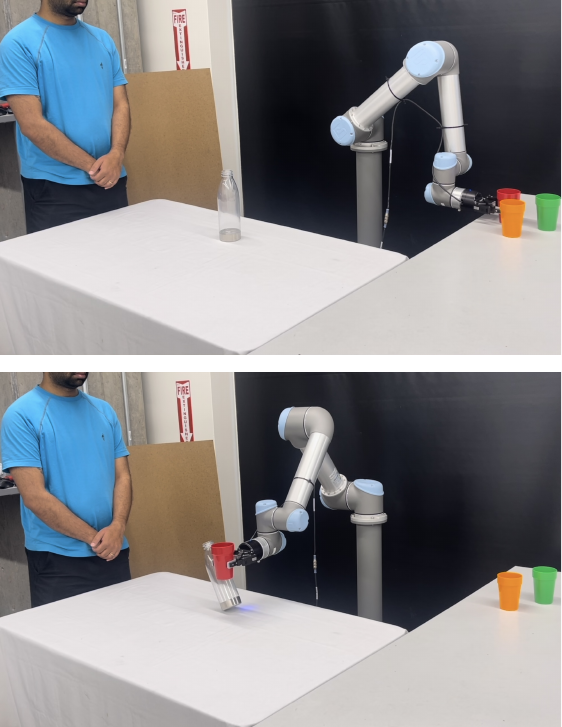}
    \label{fig:ur5-mrh}}
    \caption{Robot optimal behavior and explicable behavior.}
    \label{fig:ur5-mr-mrh}
    \vskip-5pt
\end{figure}

In this experiment, we analyzed the behaviors generated when a physical UR5 robot assisted a user to set a dining table, to demonstrate that SEPS also applies to task planning domains. The experiment (Fig. \ref{fig:ur5-mr-mrh}) was designed in a discrete setting with preset object locations, and robot motions calibrated using the UR5 inverse kinematics library. The states were defined by the locations of objects (A, B, C, D, E, shown in Fig. \ref{fig:ur5-mr}) and their status (steady or slipped).

We assumed a mismatch between robot functionality and user expectations. In $\mathcal{M}_A$, the robot is tasked with passing cups to the user by placing them anywhere on the dining table. In contrast, the user expects the cups to be placed near him. However, it risks tipping the glass in front of the robot and incurs a high cost. We evaluated SEPS against AGT, HUM, and SEPS$+\mathcal{R}_A{-}C_0$.

\begin{figure}[ht!]
    \centering
    \subfigure[$\tau_{\text{SEPS}+\mathcal{R}_A{-}C_0}$]
    {\includegraphics[scale=0.35]{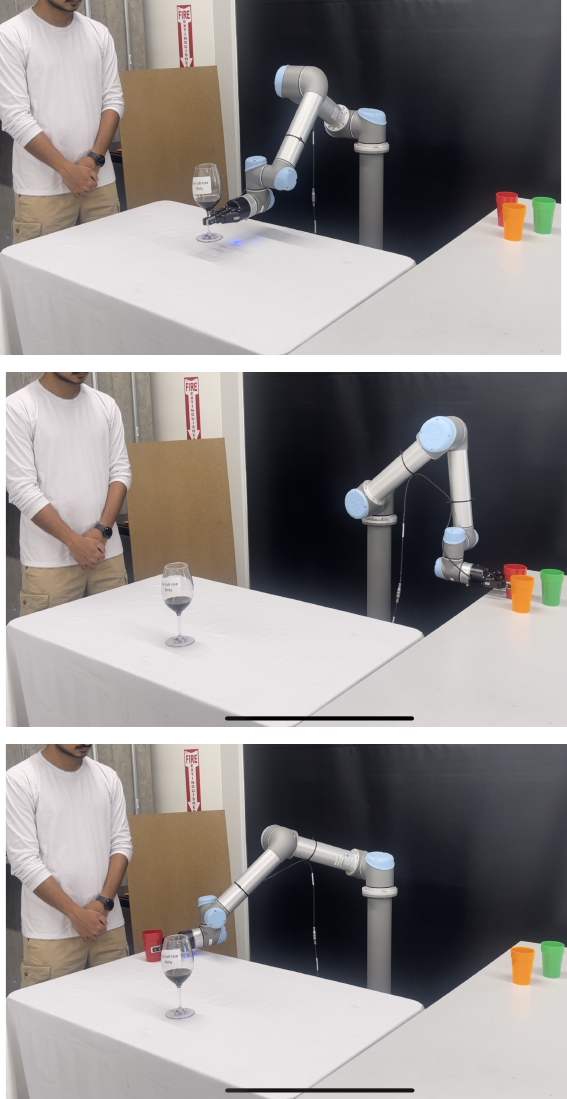}
    \label{fig:ur5-eps}}
    \hspace{20pt}
    \subfigure[$\tau_\text{SEPS}$]
    {\includegraphics[scale=0.35]{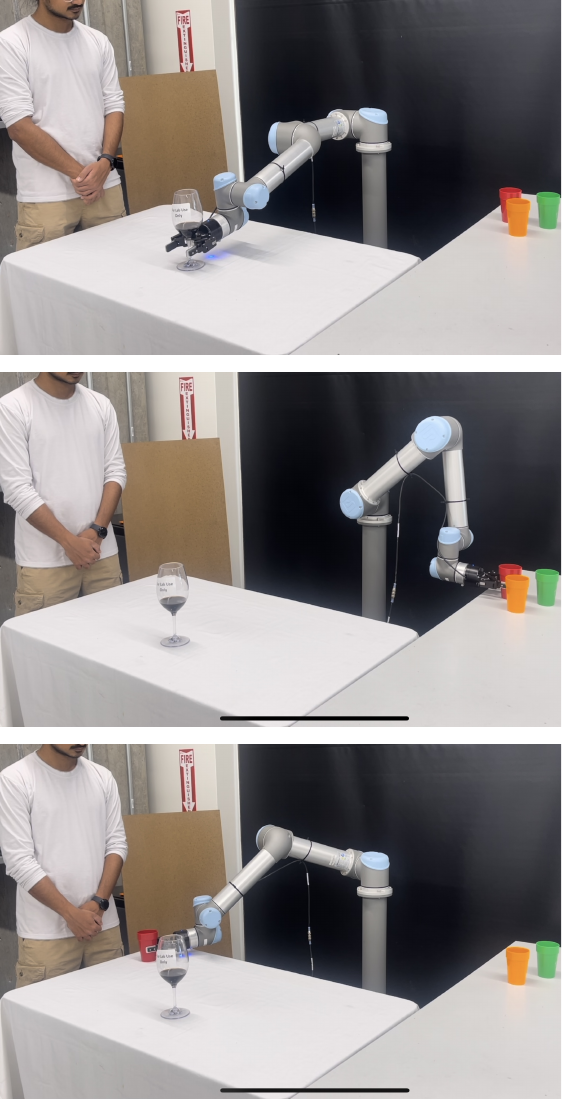}
    \label{fig:ur5-seps}}
    \caption{Safe explicable behaviors generated by SEPS$+\mathcal{R}_A{-}C_0$ and SEPS. 
    }
    \label{fig:ur5}
    \vskip-15pt
\end{figure}

\noindent\textbf{Results and Discussion.}
In AGT, the robot picked up the cup and placed it on the table next to the glass (Fig. \ref{fig:ur5-mr}). In HUM, the robot tipped the glass (replaced with an empty plastic bottle for safety), which in reality would have spilled liquid due to obstruction of the manipulator (Fig. \ref{fig:ur5-mrh}). In SEPS$+\mathcal{R}_A{-}C_0$ (Fig. \ref{fig:ur5-eps}), the robot first picked up and placed the glass out of the way before placing the cup near the user, which met the user's expectations but was inefficient due to extra movements. In SEPS (Fig. \ref{fig:ur5-seps}), the robot gently pushed the glass aside and then placed the cup near the user, achieving both explicability and efficiency. Thus, while both SEPS$+\mathcal{R}_A{-}C_0$ and SEPS produced explicable behaviors, SEPS was more efficient due to the suboptimality constraint.

\section{Conclusion}

\textcolor{black}{
In this paper, we introduced Safe Explicable Policy Search (SEPS), which addresses the problem of searching for explicable policies when both the environment model and user expectations are unknown, while ensuring safety and efficiency. We adopted the surrogate reward model from EPS as the explicability metric in SEPS, which allowed a reduction of optimization in two models to a single model. This reduction allowed us to formulate the problem as a CMDP, which simplifies the specification of safety:
the thresholds for the expected returns in CMDP are relatively straightforward to set as opposed to tuning weights and sensitivity parameters in a risk-aware RL framework.
Similarly, unlike EPS, SEPS alleviates the need for multiple hyperparameters by treating the agent’s task objective as a constraint.
The CPO method elegantly solves CMDP problems, owing to its convergence and safety guarantees, but is limited to a single constraint. We addressed this by deriving an analytical solution to SEPS for the common case with two constraints.
Our evaluation in Safety-Gym environments showed that SEPS learned safe, explicable, and efficient policies, and we further demonstrated its applicability through a physical robot experiment.
}

\textcolor{black}{
Meanwhile, there are several limitations with SEPS to consider in future work. 
The complexity of the solution increases with the number of constraints, which remains to be studied when there are more than two constraints.
Imposing both the suboptimality and safety constraints can lead to search failures. 
In such cases, feasible solutions can be obtained by relaxing the suboptimality criterion.
How to bound such relaxations is a challenging problem. 
Since CPO ensures safety only in expectation with respect to state visitations, agents may still violate safety constraints during and after learning with stricter safety criteria, such as trajectory-level constraints. Addressing this limitation, for instance, through shielding methods \cite{alshiekh2018safe}, remains an interesting direction for future work.
}     



\bibliographystyle{IEEEtran}
\bibliography{bibtex}




\section*{Supplementary material}

# Analytical solution for SEPS

## 1 Theory

The SEPS problem is approximated by

$$\begin{aligned}
\theta_{n+1} = \arg\max_\theta \; & g^T(\theta - \theta_n) \\
\text{s.t.} \; & c_0 - b_0^T(\theta - \theta_n) \leq 0 \\
& c_i + b_i^T(\theta - \theta_n) \leq 0 \quad i = 1, ..., k \\
& \frac{1}{2}(\theta - \theta_n)^T H(\theta - \theta_n) \leq \delta
\end{aligned} \quad (1)$$

Here, we derive the solution for the special case with a single cost function $\mathcal{C}_1$ which results in a total of two constraints. The resulting objective with linear and quadratic constraints is discussed under the feasible and infeasible case.

### 1.1 Feasible case

Setting $x = \theta_n - \theta$, and switching the sign of $b_i$ for $i \geqslant 1$ for convenience of notation, we are left to solve the problem in the following theorem.

**Theorem 1** *Consider the optimization problem*

$$p^* = \min_x \; g^T x$$

$$\begin{aligned}
\text{s.t.} \; & b_0^T x + c_0 \leq 0 \\
& b_1^T x + c_1 \leq 0 \\
& \frac{1}{2} x^T H x \leq \delta
\end{aligned}$$

*where $g, b_0, b_1, x \in \mathbb{R}^l$, $c_0, c_1, \delta \in \mathbb{R}$, and $H$ is a real symmetric $l \times l$-matrix. Assume that $H$ is positive definite, so that that associated quadratic form is convex. Furthermore, assume there is a strictly feasible solution to the problem (Slater condition).*

*Then the value $x^*$ corresponding to the optimal solution can expressed as*

$$x^* = -\frac{1}{\lambda^*} H^{-1}\left(g + \nu_0^* b_0 + \nu_1^* b_1\right),$$

*where*
$$\lambda^* = \arg\max_\lambda \left\{ \max_{\lambda \in \Lambda_i} f_i(\lambda),\ i = 1, 2, 3, 4 \right\},$$

with $f_1(\lambda)$, $f_2(\lambda)$, $f_3(\lambda)$, $f_4(\lambda)$, $\Lambda_1$, $\Lambda_2$, $\Lambda_3$, $\Lambda_4$, as defined below, and $\nu_0^*$, $\nu_1^*$ the corresponding values in the optimal case.

*Set*
$$q = g^T H^{-1} g, \qquad r_0 = g^T H^{-1} b_0, \qquad r_1 = g^T H^{-1} b_1,$$
$$s_0 = b_0^T H^{-1} b_0, \qquad s_1 = b_1^T H^{-1} b_1, \qquad t = b_0^T H^{-1} b_1.$$

$\underline{f_1(\lambda),\ \Lambda_1:}$

$$f_1(\lambda) = \frac{1}{2\lambda}\left(\frac{s_1 r_0^2 + s_0 r_1^2 - 2t r_0 r_1}{s_0 s_1 - t^2} - q\right) + \frac{\lambda}{2}\left(\frac{s_1 c_0^2 + s_0 c_1^2 - 2t c_0 c_1}{s_0 s_1 - t^2} - 2\delta\right) + \frac{t r_1 c_0 + t r_0 c_1 - s_1 r_0 c_0 - s_0 r_1 c_1}{s_0 s_1 - t^2}.$$

$$\Lambda_1 = \{\lambda \mid \lambda(s_0 c_1 - t c_0) > s_0 r_1 - t r_0,\ \lambda(s_1 c_0 - t c_1) > s_1 r_0 - t r_1,\ \lambda \geq 0\},$$

$$\nu_0^* = \frac{\lambda(s_1 c_0 - t c_1) + t r_1 - s_1 r_0}{s_0 s_1 - t^2}, \qquad \nu_1^* = \frac{\lambda(s_0 c_1 - t c_0) + t r_0 - s_0 r_1}{s_0 s_1 - t^2}.$$

$\underline{f_2(\lambda),\ \Lambda_2:}$

$$f_2(\lambda) = \frac{1}{2\lambda}\left(\frac{r_0^2}{s_0} - q\right) + \frac{\lambda}{2}\left(\frac{c_0^2}{s_0} - 2\delta\right) - \frac{r_0 c_0}{s_0},$$

$$\Lambda_2 = \{\lambda \mid \lambda c_0 > r_0,\ \lambda \geq 0\}.$$

$$\nu_0^* = \frac{\lambda c_0 - r_0}{s_0}, \qquad \nu_1^* = 0.$$

$\underline{f_3(\lambda),\ \Lambda_3:}$

$$f_3(\lambda) = \frac{1}{2\lambda}\left(\frac{r_1^2}{s_1} - q\right) + \frac{\lambda}{2}\left(\frac{c_1^2}{s_1} - 2\delta\right) - \frac{r_1 c_1}{s_1},$$

$$\Lambda_3 = \{\lambda \mid \lambda c_1 > r_1,\ \lambda \geq 0\}.$$

$$\nu_0^* = 0, \qquad \nu_1^* = \frac{\lambda c_1 - r_1}{s_1}.$$

$\underline{f_4(\lambda),\ \Lambda_4:}$

$$f_4(\lambda) = \frac{1}{2\lambda}(-q) + \frac{\lambda}{2}(-2\delta).$$

$$\Lambda_3 = \{\lambda \mid \lambda \geq 0\}.$$

$$\nu_0^* = 0, \qquad \nu_1^* = 0.$$

**Proof:** Since strong duality applies under the given assumptions, it suffices to solve the Lagrangian dual problem:

$$p^* = \max_{\lambda \geq 0, \nu_0 \geq 0, \nu_1 \geq 0} \min_x \mathcal{L}(x, \lambda, \nu_0, \nu_1), \quad (2)$$

where

$$\mathcal{L}(x, \lambda, \nu_0, \nu_1) \doteq g^T x + \lambda \left( \frac{1}{2} x^T H x - \delta \right) + \nu_0 \left( b_0^T x + c_0 \right) + \nu_1 \left( b_1^T x + c_1 \right)$$

From

$$\left( \frac{\partial \mathcal{L}(x, \lambda, \nu_0, \nu_1)}{\partial x} \right)^T = \lambda H x + g + \nu_0 b_0 + \nu_1 b_1 = 0$$

we obtain the critical point

$$x^* = -\frac{1}{\lambda} H^{-1} \left( g + \nu_0 b_0 + \nu_1 b_1 \right).$$

Substituting $x^*$ in Eq. (2) we have:

$$p^* = \max_{\lambda \geq 0, \nu_0 \geq 0, \nu_1 \geq 0} \ell(\lambda, v_0, v_1), \quad (3)$$

where

$$\ell(\lambda, v_0, v_1) = -\frac{1}{2\lambda} \left( q + 2\nu_0 r_0 + 2\nu_1 r_1 + \nu_0^2 s_0 + 2\nu_0 \nu_1 t + \nu_1^2 s_1 \right)$$
$$+ \nu_0 c_0 + \nu_1 c_1 - \lambda \delta, \quad (4)$$

and

$$q = g^T H^{-1} g, \qquad r_0 = g^T H^{-1} b_0, \qquad r_1 = g^T H^{-1} b_1,$$
$$s_0 = b_0^T H^{-1} b_0, \qquad s_1 = b_1^T H^{-1} b_1, \qquad t = b_0^T H^{-1} b_1.$$

From

$$\frac{\partial \ell(\lambda, \nu_0, \nu_1)}{\partial \nu_0} = -\frac{1}{\lambda} \left( r_0 + \nu_0 s_0 + \nu_1 t \right) + c_0 = 0$$

we obtain

$$\nu_0 = \frac{\lambda c_0 - \nu_1 t - r_0}{s_0}. \quad (5)$$

Likewise,

$$\frac{\partial \ell(\lambda, \nu_0, \nu_1)}{\partial \nu_1} = -\frac{1}{\lambda} \left( r_1 + \nu_1 s_1 + \nu_0 t \right) + c_1 = 0$$

leads to

$$\nu_1 = \frac{\lambda c_1 - \nu_0 t - r_1}{s_1}. \quad (6)$$

Solving Eq. (5) and Eq. (6) simultaneously, we get the critical point

$$\nu_0^* = \frac{\lambda(s_1 c_0 - t c_1) + t r_1 - s_1 r_0}{s_0 s_1 - t^2} \quad \text{and} \quad \nu_1^* = \frac{\lambda(s_0 c_1 - t c_0) + t r_0 - s_0 r_1}{s_0 s_1 - t^2}. \quad (7)$$

Here, we assume without loss of generality that $s_0 s_1 - t^2 \neq 0$, i.e., $b_0, b_1$ are not proportional. Thus, we observe that $s_0 s_1 - t^2 > 0$ by the Cauchy-Schwarz inequality.

**Case 1:** $\nu_0 > 0$ and $\nu_1 > 0$, i.e, the critical point lies in the interior of the domain.

Substitute the values of $\nu_0^*$ and $\nu_1^*$ from Eq. (7) into Eq. (4), to obtain

$$f_1(\lambda) \doteq \frac{1}{2\lambda}\left(\frac{s_1 r_0^2 + s_0 r_1^2 - 2t r_0 r_1}{s_0 s_1 - t^2} - q\right) + \frac{\lambda}{2}\left(\frac{s_1 c_0^2 + s_0 c_1^2 - 2t c_0 c_1}{s_0 s_1 - t^2} - 2\delta\right) + \frac{t r_1 c_0 + t r_0 c_1 - s_1 r_0 c_0 - s_0 r_1 c_1}{s_0 s_1 - t^2}. \quad (8)$$

Therefore, in this case we reduce to the problem

$$p^* = \max_{\lambda \in \Lambda_1} f_1(\lambda),$$

where

$$\Lambda_1 = \{\lambda \,|\, \lambda(s_0 c_1 - t c_0) > s_0 r_1 - t r_0, \;\; \lambda(s_1 c_0 - t c_1) > s_1 r_0 - t r_1, \;\; \lambda \geq 0\}.$$

We observe that

$$s_1 r_0^2 + s_0 r_1^2 - 2t r_0 r_1 - q(s_0 s_1 - t^2) \leq 0$$

as this is the negative of the Gram determinant of $g, b_0, b_1$ in the inner product associated to $H^{-1}$. Then, the coefficient of $\frac{1}{\lambda}$ in $f_1(\lambda)$ is non-positive.

Since we are in the case where both dual variables $\nu_0$ and $\nu_1$ are active, by complementary slackness of the KKT conditions the optimal solution to the original problem must satisfy the equalities $b_0^T x + c_0 = 0$ and $b_1^T x + c_1 = 0$. By the feasibility assumption, the optimal solution $d^*$ to

$$d = \min_x \frac{1}{2} x^T H x$$

subject to

$$\text{s.t.} \quad b_0^T x + c_0 = 0 \\ b_1^T x + c_1 = 0,$$

which is

$$d^* = \frac{s_1 c_0^2 + s_0 c_1^2 - 2t c_0 c_1}{2(s_0 s_1 - t^2)},$$

must satisfy $d^* \leq \delta$. This implies that the coefficient of $\frac{1}{\lambda}$ in $f_1(\lambda)$ is also non-positive.

**Case 2:** $\nu_0 > 0$ and $\nu_1 = 0$, i.e, optimizing along the "$\nu_0$-axis".

Set $\nu_1 = 0$ in Eq. (3), to get

$$p^* = \max_{\lambda \geq 0, \nu_0 \geq 0} -\frac{1}{2\lambda}\left(q + 2\nu_0 r_0 + \nu_0^2 s_0\right) + \nu_0 c_0 - \lambda \delta. \tag{9}$$

The function in Eq. (9) has critical point

$$\nu_0 = \frac{\lambda c_0 - r_0}{s_0}.$$

Substitute the value of $\nu_0$ above in Eq. (9), to reduce to the problem

$$p^* = \max_{\lambda \in \Lambda_2} f_2(\lambda) \doteq \frac{1}{2\lambda}\left(\frac{r_0^2}{s_0} - q\right) + \frac{\lambda}{2}\left(\frac{c_0^2}{s_0} - 2\delta\right) - \frac{r_0 c_0}{s_0},$$

where

$$\Lambda_2 = \{\lambda \,|\, \lambda c_0 > r_0,\ \lambda \geq 0\}.$$

Similarly as in Case 1, we have that the coefficients of $\lambda$ and $\frac{1}{\lambda}$ in $f_2(\lambda)$ are non-positive.

**Case 3:** $\nu_0 = 0$ *and* $\nu_1 > 0$, *i.e, optimizing along the "$\nu_1$-axis"*.

This case is analogous to the previous one. Thus we reduce to the problem

$$p^* = \max_{\lambda \in \Lambda_3} f_3(\lambda) \doteq \frac{1}{2\lambda}\left(\frac{r_1^2}{s_1} - q\right) + \frac{\lambda}{2}\left(\frac{c_1^2}{s_1} - 2\delta\right) - \frac{r_1 c_1}{s_1},$$

where

$$\Lambda_3 = \{\lambda \,|\, \lambda c_1 > r_1,\ \lambda \geq 0\}.$$

Similarly as in Case 1, we have that the coefficients of $\lambda$ and $\frac{1}{\lambda}$ in $f_3(\lambda)$ are non-positive.

**Case 4:** $\nu_0 = 0$ *and* $\nu_1 = 0$.

Set $\nu_0 = 0$ and $\nu_1 = 0$ in Eq. (3), to get

$$p^* = \max_{\lambda \geq 0} f_4(\lambda) \doteq \frac{1}{2\lambda}(-q) + \frac{\lambda}{2}(-2\delta). \tag{10}$$

We note that the coefficients of $\lambda$ and $\frac{1}{\lambda}$ in $f_4(\lambda)$ are non-positive.

In summary, in each of the four cases we reduced the problem to solving a problem

$$\max_{\lambda \in \Lambda} \frac{1}{2}\left(\frac{A}{\lambda} + B\lambda\right) + C \tag{11}$$

for some $A \leq 0$ and $B \leq 0$, and $\Lambda \subset \mathbb{R}$ a certain interval. This can be easily solved as a single-variable optimization problem.

If $\lambda^*$, $\nu_0^*$, and $\nu_1^*$ give an optimal solution to the dual problem in Theorem 1, the update rule for the feasible case is (using the original parameters)

$$\theta_{n+1} = \theta_n + \frac{1}{\lambda^*} H^{-1}\left(g + \nu_0^* b_0 - \nu_1^* b_1\right). \tag{12}$$

## 1.2 Infeasible case

In the infeasible case, we set $x = \theta - \theta_n$, and switch the sign of $m = 0$ and keep the sign as is for $m \geq 1$, to solve for the general violation case $c_m + b_m^T x$.

**Theorem 2** *Consider the optimization problem*

$$\hat{p} = \min_x \ c_m + b_m^T x$$

$$s.t. \ \frac{1}{2} x^T H x \leq \delta$$

*where $b_m, x \in \mathbb{R}^l, c_m, \delta \in \mathbb{R}$, and $H$ is a real symmetric $l \times l$-matrix. Assume that $H$ is positive definite, so that that associated quadratic form is convex. Furthermore, assume there is a strictly feasible solution to the problem (Slater condition).*

*Then the value $x^*$ corresponding to the optimal solution can expressed as*

$$x^* = -\frac{1}{\phi^*} H^{-1} b,$$

*where $\phi^*$ is defined by*

$$\phi^* = \sqrt{\frac{b^T H^{-1} b}{2\delta}}.$$

**Proof:** Since strong duality applies under the given assumptions, it suffices to solve the Lagrangian dual problem:

$$\hat{p} = \max_{\phi \geq 0} \min_x \ \mathcal{L}(x, \phi) \doteq \frac{\phi}{2} x^T H x + b_m^T x + c_m - \phi \delta \tag{13}$$

From

$$\left(\frac{\partial \mathcal{L}(x, \phi)}{\partial x}\right)^T = \phi H x + b_m = 0$$

we obtain the critical point

$$x^* = -\frac{1}{\phi} H^{-1} b_m$$

Substituting $x^*$ in Eq. (13) we have:

$$\hat{p} = \max_{\phi \geq 0} \ \ell(\phi) \doteq -\frac{s}{2\phi} + c_m - \phi \delta, \quad \text{where} \quad s = b^T H^{-1} b.$$

Finally, from

$$\frac{\partial \ell(\phi)}{\partial \phi} = \frac{s}{2\phi^2} - \delta = 0$$

we obtain the critical point

$$\phi^* = \sqrt{\frac{s}{2\delta}}.$$

If $\phi^*$ is the solution to the dual above. The update rule for the infeasible case is

$$\theta_{n+1} = \theta_n - \sqrt{\frac{2\delta}{s}} H^{-1} b_m \tag{14}$$

6



\end{document}